\documentclass{article}

\usepackage[utf8]{inputenc} 
\usepackage[T1]{fontenc}    
\usepackage{hyperref}       
\usepackage{url}            
\usepackage{xfrac}
\usepackage{booktabs}       
\usepackage{amsfonts}       
\usepackage{nicefrac}       
\usepackage{microtype}      
\usepackage{graphicx}
\usepackage{amsmath}
\usepackage{color}
\usepackage[dvipsnames]{xcolor}
\usepackage{comment}
\usepackage{multirow}
\usepackage{pifont}
\usepackage{threeparttable}
\usepackage{colortbl}

\usepackage{mathtools}
\usepackage{amssymb}
\usepackage{latexsym}
\usepackage{dsfont}
\usepackage{mathrsfs}
\usepackage{amssymb}
\usepackage{amsfonts}
\usepackage{bm}
\usepackage{xspace}
\usepackage{amsthm}

\ifx\BlackBox\undefined
\newcommand{\BlackBox}{\rule{1.5ex}{1.5ex}}  
\fi

\ifx\QED\undefined
\def\QED{~\rule[-1pt]{5pt}{5pt}\par\medskip}
\fi

\ifx\proof\undefined
\newenvironment{proof}{\par\noindent{\em Proof:\ }}{\hfill\BlackBox\\[.0mm]}
\fi

\ifx\theorem\undefined
\newtheorem{theorem}{Theorem}[section]
\fi

\ifx\example\undefined
\newtheorem{example}{Example}[section]
\fi

\ifx\property\undefined
\newtheorem{property}{Property}
\fi

\ifx\lemma\undefined
\newtheorem{lemma}{Lemma}[section]
\fi

\ifx\proposition\undefined
\newtheorem{proposition}{Proposition}[section]
\fi

\ifx\remark\undefined
\newtheorem{remark}{Remark}
\fi

\ifx\corollary\undefined
\newtheorem{corollary}{Corollary}[section]
\fi

\ifx\definition\undefined
\newtheorem{definition}{Definition}[section]
\fi

\ifx\conjecture\undefined
\newtheorem{conjecture}{Conjecture}[section]
\fi

\ifx\axiom\undefined
\newtheorem{axiom}[theorem]{Axiom}
\fi

\ifx\claim\undefined
\newtheorem{claim}[theorem]{Claim}
\fi

\ifx\assumption\undefined
\newtheorem{assumption}{Assumption}
\fi

\ifx\problem\undefined
\newtheorem{problem}{Problem}[section]
\fi

\ifx\fact\undefined
\newtheorem{fact}{Fact}
\fi

\newlength\savewidth\newcommand\shline{\noalign{\global\savewidth\arrayrulewidth
  \global\arrayrulewidth 1pt}\hline\noalign{\global\arrayrulewidth\savewidth}}
\newcommand{\tablestyle}[2]{\setlength{\tabcolsep}{#1}\renewcommand{\arraystretch}{#2}\centering\footnotesize}

\definecolor{dkgreen}{rgb}{0,0.6,0}
\definecolor{gray}{rgb}{0.5,0.5,0.5}
\definecolor{mauve}{rgb}{0.58,0,0.82}
\definecolor{GrayBG}{gray}{0.9}

\usepackage[accepted]{icml2021}

\icmltitlerunning{Benchmarking Graphormer \\ on Large-Scale Molecular Modeling Datasets}

\begin{document}

\twocolumn[
\icmltitle{Benchmarking Graphormer \\ on Large-Scale Molecular Modeling Datasets}



\icmlsetsymbol{equal}{*}

\author{%
Yu Shi$^1$, \ Shuxin Zheng$^1$\thanks{Contact Person.},\ \ Guolin Ke$^1$, \\ \textbf{Yifei Shen}$^2$\ \, \textbf{Jiacheng You}$^3$\thanks{Interns at MSRA},\ \ \textbf{Jiyan He}$^4$\footnotemark[2],\ \ \textbf{Shengjie Luo}$^5$\footnotemark[2],\ \ \textbf{Di He}$^1$,\  \textbf{Tie-Yan Liu}$^1$\\
$^1$Microsoft Research Asia \quad $^2$ HKUST \\
$^3$ Tsinghua University \quad $^4$ USTC \quad $^5$ Peking University \\
\texttt{\footnotesize shuz@microsoft.com} \\
}

\begin{icmlauthorlist}
\icmlauthor{Yu Shi}{msra}
\icmlauthor{Shuxin Zheng}{msra}
\icmlauthor{Guolin Ke}{msra}
\icmlauthor{Yifei Shen}{hkust}
\icmlauthor{Jiacheng You}{thu}
\icmlauthor{Jiyan He}{ustc}
\icmlauthor{Shengjie Luo}{pku}
\icmlauthor{Chang Liu}{msra}
\icmlauthor{Di He}{msra}
\icmlauthor{Tie-Yan Liu}{msra}
\end{icmlauthorlist}

\icmlaffiliation{ustc}{USTC}
\icmlaffiliation{hkust}{HKUST}
\icmlaffiliation{thu}{Tsinghua University}
\icmlaffiliation{pku}{Peking University}
\icmlaffiliation{msra}{Microsoft Research Asia}

\icmlcorrespondingauthor{Shuxin Zheng}{shuz@microsoft.com}

\icmlkeywords{Graphormer, Molecular Modeling}

\vskip 0.3in
]



\printAffiliationsAndNotice{}  

\begin{abstract}
This technical note describes the recent updates of Graphormer, including architecture design modifications, and the adaption to 3D molecules. With these simple modifications, Graphormer attained better results on large-scale molecular modeling datasets, and the performance gain could be consistently obtained on 2D and 3D molecular graph modeling tasks. In addition, we show that with a global receptive field and an adaptive aggregation strategy, Graphormer is more powerful than classic message-passing-based GNNs. Empirically, Graphormer could achieve much less MAE than the originally reported results on the PCQM4M quantum chemistry dataset used in KDD Cup 2021. In the meanwhile, it greatly outperforms the competitors in the recent Open Catalyst Challenge, which is a competition track on NeurIPS 2021 workshop, and aims to model the catalyst-adsorbate reaction system with advanced AI models. All code can be found at this \href{https://github.com/Microsoft/Graphormer}{link}.

\end{abstract}

\section{Introduction}
Graphormer~\cite{ying2021transformers} is a recently proposed deep learning model built upon the standard Transformer, which aims to break through the limitations of conventional graph neural networks~\cite{xu2018how, hamilton2017inductive} on expressiveness and over-smoothing. The architecture design of Graphormer is quite simple, where a standard Transformer~\citep{vaswani2017attention} is equipped by three structural encodings (i.e., centrality, spatial, and edge encodings), 
which is attractively effective on a wide range of graph representation tasks. Graphormer enjoys the great power of expressiveness from Transformer architecture, while also incorporates the structural information of the graph with high efficiency.

In this note, we report several design improvements built in the Graphormer framework. We first investigate the change of placement of layer normalization, and find that the Post-LN variant could lead to significantly better results on molecular property prediction tasks. Furthermore, we extend Graphormer to 3D molecule graph modeling by designing specific centrality encoding, spatial encoding, and 3D attention sub-layer. After a series of architecture design modifications, the upgraded Graphormer establishes stronger and more feasible baselines on large-scale molecular modeling datasets, i.e., PCQM4M~\cite{hu2021ogb} and OC20~\cite{chanussot2020open}. 

Finally, we endeavor to obtain a better theoretical understanding of Graphormer via the lens of distributed computing theory. Specifically, we show that with a global receptive field, Graphormer enjoys a greater expressiveness compared to classic message-passing-based GNNs. 

\section{Empirical Analysis}

\subsection{The Placement of Layer Normalization Matters}

The originally designed Graphormer follows the architecture of a prevailing variant of Transformer, i.e., the Pre-LN Transformer, where layer normalization is placed inside the residual blocks. Recent literature~\cite{xiong2020layer} shows that with this modification, the gradients are well-behaved at initialization, leading to a much faster convergence compared to the vanilla Post-LN Transformer~\cite{vaswani2017attention}. Yet, we observe that the latter leads to a better generalization performance on the large-scale quantum chemical property prediction dataset PCQM4M~\cite{hu2021ogb}. Therefore, we adopt this simple modification to Graphormer and establish a stronger baseline built in the Graphormer framework.

\subsubsection{Molecular Property Prediction}
\paragraph{Datasets.} The experiment is conducted on the large-scale molecular graph dataset PCQM4M~\cite{hu2021ogb}, which contains 3.8M graphs and 55.4M edges in total. It is a quantum chemistry dataset aiming to accelerate quantum physical property calculation based on the Density Functional Theory (DFT) by advanced machine learning methods. This dataset has been recently updated to v2 with several modifications\footnote{\url{https://ogb.stanford.edu/docs/lsc/pcqm4mv2/}.} and 3D molecular structures. In this section, we report the performance on both versions, but without using any 3D geometric information. 

\paragraph{Settings.} We maintain the model configuration for the 12-layer Graphormer$_{\small Base}$ model, then we further scale up it to a 24-layer Graphormer$_{\small Large}$ model:
\begin{enumerate}
    \item Graphormer$_{\small Base}$: $L=12, d=768, H=32$,
    \item Graphormer$_{\small Large}$: $L=24, d=1024, H=32$,
\end{enumerate}
where $d$ and $H$ represents the hidden dimension and the number of attention heads, respectively. We keep the hyper-parameter the same as~\cite{ying2021transformers}, except that a small learning rate $8\times 10^{-5}$ is used for PostLN$_{\small{Large}}$ to improve training stability.

\begin{table}[t]
\vspace{0.5em}
\begin{center}
\tablestyle{5.5pt}{1.0}
\begin{tabular}{l|c|cc|cc}
~  & ~ & \multicolumn{2}{c|}{\scriptsize{MAE on PCQv1}} & \multicolumn{2}{c}{\scriptsize{MAE on PCQv2}} \\
case & \#params. & train. & valid. &  train. & valid. \\
\shline
PreLN$_{\small{Base}}$ & 48.3M & 0.0266 & 0.1229 & 0.0266 & 0.0889 \\
PreLN$_{\small{Large}}$ & 159.3M & 0.0172 & 0.1213 & 0.0173 & 0.0879 \\
\rowcolor{GrayBG} PostLN$_{\small{Base}}$ & 48.3M & 0.0416 & \textbf{0.1193} & 0.0348 & \textbf{0.0864} \\
PostLN$_{\small{Large}}$ & 159.3M & 0.0212 & 0.1228 & 0.0186 & 0.0883 \\

\end{tabular}
\end{center}
\caption{\textbf{Graphormer's PreLN variant vs. PostLN variant}: Mean absolute error on PCQM4M dataset. Our new Graphormer implementation is based on \textit{fairseq}\footnote{\url{https://github.com/pytorch/fairseq/}.}, and may lead to slightly different numbers with ~\citet{ying2021transformers}}
\label{tab:pcq}
\end{table}

From Table \ref{tab:pcq}, we can see the PostLN variant could attain better performance than PreLN on the large-scale molecular property prediction task. Interestingly, although the optimization error is lower, the generalization ability of Graphormer$_{\small Large}$ is worse on this dataset. Thus, there is still potential to develop deep Graphormer models with better generalization.

\begin{table*}[t]
\vspace{0.5em}
\begin{center}
\begin{tabular}{l|c|c|c|c|c}
~  &  \multicolumn{5}{c}{\scriptsize{Enery MAE (eV) on IS2RE Task (Direct)}}  \\
case & ID & OOD Ads. & OOD Cat. &  OOD Both & avg. \\
\shline
Graphormer$_{\small{Base}}$\scriptsize{*}    & 0.4329  & 0.5850  & 0.4441    &  0.5299 & 0.4980 \\
Graphormer$_{\small{Base}}$ (\textit{ensemble})  & 0.3976  & 0.5719    & 0.4166    & 0.5029    &  0.4722   \\

\end{tabular}
\end{center}
\caption{Results on IS2RE task by direct approach. {\scriptsize{*}} denotes evaluation on the OC20 validation split.}\label{tab:3}
\end{table*}

\subsection{Adaptation to 3D Molecular Modeling}

A molecule can be represented by a 3D molecular graph $G=(V,P)$, where $V=\{v_1, v_2, \cdots, v_n\}$ denotes the set of atoms, each of which holds a feature vector $x_i$, and $P=\{r_1, r_2, \cdots, r_n\}$ is the set of 3D Cartesian coordinates of atoms which contains 3D spatial information. Given a 3D molecular graph as the input of Graphormer, new designs of structural encoding (i.e., spatial encoding and centrality encoding in ~\citet{ying2021transformers}) are desired for modeling the graph spatial information. First, we choose $\phi(v_i,v_j)$ in the spatial encoding to be the Euclidean distance between $v_i$ and $v_j$, and adopt a set of Gaussian basis functions~\cite{shuaibi2021rotation} to encode $\phi(v_i,v_j)$ in order to model the spatial relation between atoms. Second, all spatial encodings of each node are simply summed up to obtain the centrality encoding which describes the importance of the atom in the 3D molecular graph. 

Periodic boundary conditions (PBC) is common for crystal systems, where a set of atoms in the 3D unit cell is periodically repeated. Typically, a radius graph with PBC is constructed~\cite{zitnick2020introduction} to capture the local 3D structure surrounding each atom, where the replicated atoms among different unit cells are reduced to a single atom, may result in multiple edges between two atoms (i.e. multigraph). Since message passing is done by attention layers in Graphormer, rather than constructing a multigraph, we prefer to simply physically duplicate all atoms if they lie within the cutoff distance in multiple repeated cells.

In addition, we design a new attention layer to replace the original node-level projection head for generating 3D outputs. Concretely, the attention probability in a standard self-attention layer is decomposed into three directions by multiplying the normalized relative position offset $\frac{\textbf{r}_{ij}}{\left\Vert \textbf{r}_{ij} \right\Vert}\in \mathbb{R}^3$ between query and key atoms. Then three linear projection heads are applied to each component of the 3D attention layer's output in the three directions respectively. Specifically, it could maintain rotational equivariance of the model's final estimation layer if the parameters of the three linear projections are shared.

\subsubsection{Electrocatalyst}

\paragraph{Dataset.} We verify the effectiveness of 3D molecular modeling on the recent electrocatalysts dataset - the Open Catalyst 2020 (OC20)~\cite{chanussot2020open}. It aims to accelerate the catalyst discovery process for solar fuels synthesis, long-term energy storage, and renewable fertilizer production, by using machine learning models to find low-cost electrocatalysts to drive the electrochemical reactions at high rates. The OC20 dataset contains more than 660k catalyst-adsorbate reaction systems (over 140M structure-energy estimation) produced by molecular dynamics simulation using density functional theory. In this section, we report the results of Graphormer for predicting the relaxed energies from initial structures.

\paragraph{Settings.} We employ a 12-layer Graphormer$_{\small {Base}}$ as the basic model for energy prediction. Inspired by ~\citet{jumper2021highly}, we repeatedly feed the outputs to this basic model by four times, which contributes markedly to accuracy with minor extra training time. We optimize the model using Adam with learning rate 3e-4 and weight decay 1e-3. We train the model using batch size 64 for 1 million steps. 

In addition to predicting the relaxed energy of the entire system, inspired by ~\citet{godwin2021very}, we further adopt an auxiliary node-level objective to predict the displacement of each atom between the initial and relaxed structures. In Table~\ref{tab:3} we report the performance on the IS2RE Direct track, which directly estimates the relaxed energy from the initial structure. As shown in the table, the energy prediction of unseen element compositions for catalysts (Out of Domain (OOD) Catalyst) is much accurate than OOD Adsorbates, and OOD Both, which implies that Graphormer may have the potential to help the catalyst discovery process for well-known but important chemical species involved in the chemical reactions of interest, such as $\mathrm{OH}$, $\mathrm{O_2}$, or $\mathrm{H_2O}$.

\section{Understanding Graphormer from a Theoretical Perspective}
In this section, we develop a better theoretical understanding of Graphormer based on distributed computing theory. Compared with the classic message-passing-based GNNs (MPGNNs), Graphormer enjoys two unique characteristics: a global receptive field and an adaptive aggregation strategy. We first analyze the impact of the global receptive field on expressiveness. In literature, the $k$-Weisfeiler-Lehman ($k$-WL) graph isomorphism test is often adopted as a metric to characterize the expressiveness of MPGNNs. For example, it was shown in  \cite{xu2018how} that the anonymous MPGNN is \emph{equivalent} to $1$-WL test, which implies that anonymous MPGNN has a very limited ability in graph isomorphism test. Higher-order GNNs have been developed to go beyond $1$-WL test \cite{morris2019weisfeiler,balcilar2021breaking} and achieve the same power as $3$-WL test. In these papers, one essential assumption is that the nodes share the same feature. For wide and deep MPGNNs with discriminative features, their expressive power goes beyond $k$-WL (for any $k$) and becomes universal \cite{abboud2020surprising}. The discriminative features can be easily achieved by adding random features \cite{sato2021random} or unique identifiers \cite{loukas2019graph}, which is often adopted in practice. Thus, to characterize the expressiveness of more practical MPGNNs, a new metric is needed.

In \cite{loukas2019graph}, the expressiveness of GNNs with discriminative features was studied via the metrics in distributed computing theory. Specifically, it was proved that a $d$ layer MPGNN with width $w$ has \emph{equivalent} expressiveness to a CONGEST model with $d$ communication round and $w \log n$ communication bits in each round, where $n$ is the number of nodes in the graph. Thus, the expressiveness bound of CONGEST models in distributed computing literature can be used for characterizing the expressiveness of MPGNNs. Due to the locality of the message-passing scheme in CONGEST models, they lose a significant power when $d$ and $w$ is limited. Specifically, several graph problems cannot be solved unless the product of $d$ and $w$, i.e., communication complexity, is large enough. By repurposing these results in CONGEST models, it was shown in \cite{loukas2019graph} that several graph problems cannot be solved unless the product of MPGNN's depth $d$ and width $w$, i.e., model capacity, exceeds a polynomial of the graph size. The impossibility results in \cite{loukas2019graph} are summarized in the column ``Local MP'' in Table \ref{tab:congest}.

In contrast to the local message-passing in MPGNNs, each node is allowed to send different messages to 
any other node in Graphormer. This paradigm is called \emph{CONGESTED CLIQUE} in distributed computing literature \cite{drucker2014power} and breaks the expressiveness barrier of the CONGEST model. We summarize the corresponding possibility results of CONGESTED CLIQUE in the column ``Non-local MP'' of Table~\ref{tab:congest}. Detailed theorems are listed in Appendix~\ref{sec:cc}. These bounds demonstrate that a global receptive field qualitatively improves the expressiveness of the model.

\begin{table*}[]
\centering
\label{tab:congest}
\resizebox{0.8\textwidth}{!}{%
\begin{tabular}{@{}llll@{}}
\toprule
\textit{Problem}       & \textit{Local MP}                         & \textit{Non-local MP}                                                    \\ \midrule
4-cycle detection & $dw = \Omega({\sqrt{n}}/{\log n})$     & $dw = O(1)$ \\
subgraph verification & $d\sqrt{w} = \Omega(\sqrt{n}/\log n)$  & $dw = O(1)$\\

diam. $\sfrac{3\hspace{-1px}}{2}$-approx.      & $dw = \Omega(\sqrt{n}/{\log n})$ & $dw = O(n^{0.158})$\\

5-cycle detection  & $dw = \Omega({n}/{\log n})$            &   $dw = O(n^{0.158})$         \\
diam. computation      & $dw = \Omega({n}/{\log n})$ & $dw = O(n^{1/3}\log n)$\\
min. vertex cover       & $dw = \Omega({n^2}/{\log^2 n})$ for $w = O(1)$ & $dw = O(n^{1-2/n})$ for $w = O(1)$\\
max. indep. set       & $dw = \Omega({n^2}/{\log^2 n})$ for $w = O(1)$ & $dw = O(n^{1-2/n})$ for $w = O(1)$\\\bottomrule
\end{tabular}%
}
\caption{A comparison of local message passing and non-local message passing in the model capacity for solving different graph problems. Results for local MP are from~\cite{loukas2019graph} and results for non-local MP are shown in Appendix~\ref{sec:cc}.}
\end{table*}

\small

\bibliography{ref}
\bibliographystyle{icml2021}


\newpage
\appendix
\section{Theorems for Table \ref{tab:congest}}\label{sec:cc}
\begin{theorem} (4-cycle detection) (Theorem 4 in \cite{censor2019algebraic}) Given $w = O(1)$, the existance of 4-cycle can be detected in $O(1)$ rounds.
\end{theorem}

\begin{theorem} (5-cycle detection) (Theorem 3 in \cite{censor2019algebraic}) Given $w = O(1)$, for directed and undirected graphs, the existence of k cycles can be detected in $d = 2^{O(k)}n^{1-2/\omega}\log n$ rounds, where $\omega$ is the matrix multiplication time.
\end{theorem}
For $5$-cycle detection problem, taking $k=5$ results in the desired bound.

\begin{theorem} (Diameter computation) (Corollary 6 in \cite{censor2019algebraic}) Given $w = O(1)$, for weighed and directed graphs with integer weights $\{0, \pm 1, \cdots, \pm M\}$, all-pair shortest path can be computed in $d = O(n^{1/3}\log n \lceil \log M/\log n \rceil)$.
\end{theorem}

\begin{theorem} (Diameter approximation) (Theorem 9 in \cite{censor2019algebraic}) Given $w = O(1)$, for directed graphs with integer weights in $\{0,1, \cdots, 2^{n^{o(1)}}\}$, we can compute $(1+o(1))$-approximate all-pairs shortest path in $d = O(n^{1-2/\omega + o(1)})$ rounds.
\end{theorem}

\begin{theorem} (Subgraph verification) (Corollary 1 in \cite{jurdzinski2018mst}) Given $w = O(1)$, there are randomized distributed algorithms that solve the following verification problems in congested clique model in $O(1)$ rounds with high probability: bipartiteness verifcation, cut verification, s-t connectivity, and cycle containment.
\end{theorem}

\begin{theorem} (Subgraph Detection) \cite{doleV2012tri} Given $w = O(1)$, there are distributed algorithms that count the number of $d$-vertex subgraph in $d = O(n^{(d-2)/d}/\log n)$ rounds.
\end{theorem}
The bound of the maximum independent set can be obtained by using the relationship between $k$-independent set and maximum independent set, and the minimum vertex cover is equivalent to the maximum independent set.

\end{document}